\documentclass{llncs}
\usepackage[T1]{fontenc}
\usepackage{graphicx}
\usepackage{url} 

\title{Quality versus speed in energy demand prediction for district heating systems} 

\author{Witold Andrzejewski\inst{1}\orcidID{0000-0001-9486-929X}
\and J\k{e}drzej Potoniec\inst{1}\orcidID{0000-0002-6115-6485}
\and Maciej Drozdowski\inst{1}\orcidID{0000-0001-9314-529X}\thanks{Corresponding author}
\and Jerzy Stefanowski\inst{1}\orcidID{0000-0002-4949-8271}
\and Robert Wrembel\inst{1}\orcidID{0000-0001-6037-5718}
\and Paweł Stapf\inst{2}
}
\authorrunning{\authorrunning{W. Andrzejewski, J. Potoniec, M. Drozdowski, et al.}}
\institute{Poznan University of Technology, Pozna\'n, Poland\\
\email{\{Witold.Andrzejewski,
Jedrzej.Potoniec,
Maciej.Drozdowski,
Jerzy.Stefanowski,
Robert.Wrembel\}@cs.put.poznan.pl}
\and 
Kogeneracja Zachód S.A., Pozna\'n, Poland\\
\email{p.stapf@kogeneracjazachod.pl}
}

\begin{document}

\maketitle


\begin{abstract}
In this paper, we consider energy demand prediction in district heating systems. 
Effective energy demand prediction is essential in combined heat power systems when offering electrical energy in competitive electricity markets. 
To address this problem, we propose two sets of algorithms: (1) a novel extension to the algorithm proposed by E. Dotzauer and (2) an autoregressive predictor based on hour-of-week adjusted linear regression on moving averages of energy consumption. 
These two methods are compared against state-of-the-art artificial neural networks.
Energy demand predictor algorithms have various computational
costs and prediction quality. 
While prediction quality is a widely used measure of predictor superiority, computational costs are less frequently analyzed and their impact is not so extensively studied. 
When predictor
algorithms are constantly updated using new data, some computationally expensive forecasting methods may become inapplicable. 
The computational costs can be split into training and execution parts. 
The execution part is the cost paid when the already trained algorithm is applied to predict something.
In this paper, we evaluate the above methods with respect to the quality and computational costs, both in the training and in the execution.
The comparison is conducted on a real-world dataset from a district heating system in the northwest part of Poland.


\end{abstract}

\section{Introduction} \label{sec:Intro}
District heating systems (DHS) are widely used in North-Western Europe to deliver
heat and hot water to households. A DHS is often a cogeneration system or combined heat power (CHP), when heat is also used to generate electric energy. The amount of electric  energy that can be produced is directly related to the amount of produced heat. Moreover, it is a legal demand in Poland that district heating systems sell only electric energy cogenerated with heat, i.e., it is forbidden to produce electricity and dispose of heat. Hence, \textbf{for CHP systems, prediction of heat demand is essential} to submit bids for electric energy in the priciest hours in an energy market.

Constructing computationally efficient intelligent systems faces a number of challenges in industrial applications in which data is collected in extensive sensor networks. Often, such raw data is incomplete and of low quality. 
In this paper, we report on our experience from an R\&D project in designing algorithms for heat demand prediction for Kogeneracja Zach\'od\footnote{\url{https://kogeneracjazachod.pl/}}, a company running a cogeneration system in Poland. We evaluate the algorithms w.r.t. prediction quality and computational cost, both in the training and testing phases. The trade-off between prediction error and computational cost is interesting from the research point of view. Yet, it is essential in the edge/fog systems, if IoT devices (e.g. energy counters) with low computational resources were supposed to not only measure but also predict energy consumption and incrementally train models with new arriving data. This is a potential use-case for future smart grids, when IoT devices could optimize energy consumption plan of the consumer and energy supply to the grid.

Two types of algorithms for predicting thermal energy usage are proposed. The first one includes novel extensions of the algorithm originally proposed by E. Dotzauer in \cite{Dotz02}. The second type is based on the hour-of-week adjusted linear regression on moving averages of energy consumption and is a type of autoregressive predictor. These algorithms are compared against two state-of-the art artificial neural networks. The evaluation is conducted on real-world data which is made public for research purposes, cf. \cite{KZData}.

This paper is organized as follows. Section \ref{sec:lietrature} outlines existing literature on predicting energy consumption. In Section \ref{sec:formul} the goals in energy prediction of the R\&D project are formulated. The prediction algorithms we contribute are presented in Section \ref{sec:alg}. The test dataset is outlined in Section \ref{sec:dataset}. The prediction algorithms are evaluated in Section \ref{sec:eval}. Section \ref{sec:Concl} concludes the paper.

\section{Related Work} \label{sec:lietrature}
The most frequently used methods on predicting energy consumption, both electrical and thermal, can be roughly classified as: (1) regression, among them linear or piecewise linear methods, (2) statistical, e.g., based on autoregressive methods, (3) machine learning, e.g., neural networks, and (4) mixed ones. 

In \cite{KF11}, 12 linear and Gaussian process regression models of energy consumption were developed for residential and commercial buildings. Gaussian mixtures were also applied in \cite{MLSWYS14} to fit two-dimensional distributions of heat demand and outdoor temperature. In \cite{Dotz02} the heat demand was split into components depending on outdoor temperature and a seasonal inhabitant behavior, cf. Section \ref{sec:Dotzauer}. Heat demand for two localities in Czechia were predicted using a similar method \cite{ChrBr10}. The social component was represented by Seasonal Autoregressive Integrated Moving Average (SARIMA) model.

A Support Vector Machine model was build in \cite{DCL05} to predict a building monthly energy usage. Input parameters included monthly mean outdoor temperature, relative humidity, and global solar radiation. The model outputs an aggregated monthly prediction of energy usage. 
In \cite{ENP12}, 7 machine learning techniques predicting electrical energy consumption in residential buildings are compared: linear regression, Feed Forward Neural Networks (FFNN), Support Vector Regression, Least Squares Support Vector Machines (LS-SVM), Hierarchical Mixture of Experts (HME), HME-FFNN, Fuzzy C-Means with FFNN. The experiments were run on synthetic datasets, showing that LS-SVM offered a better prediction model. 

An FFNN with 1 hidden layer was used in \cite{BMHS08} to predict heat demand of households. In \cite{KCBR16} a Radial-Basis Function Neural Network (RBFNN) was used to model energy consumption in a large building. The goal was to minimize prediction errors in the training and validation datasets, as well as, to  minimize the RBFNN size. A hybrid genetic algorithm-adaptive network-based fuzzy inference system was proposed in \cite{LSC11}, which was a 5-layer rule-based structure. The first 3 layers calculated the strength of the rule firings while the other 2 layers calculated a weighted sum of linear functions of the input parameters. RBFNN was used also to predict heat demand on a university campus \cite{WoK14}. Conversely, to represent energy demand on another campus \cite{JSZ15}, a 3-layer FFNN, an RBFNN, and the Adaptive Neuro-Fuzzy Inference System were used.

In other studies on a DHS in Sweden, 3 methods were compared in \cite{RCCA15}: linear regression, support vector regression (SVR), and context vector regression (CVR). 
Energy demand for DHS in Riga was analyzed in \cite{PBSS17}, based on a nonlinear autoregressive neural network with external factors fed with data on the historical energy consumption and outdoor temperature. A hybrid statistical-machine learning method was proposed in \cite{XGM18}. It combined 3 methods. A chain of Seasonal Trend decomposition using LOESS \cite{CCMT90} with different season lengths was used to model heat demand. Its output, together with weather data were fed to Elman neural network. Finally, the ARIMA model was used to represent the residuum from seasonal decomposition.

We end this section with some \textbf{observations}. First, comparing the heat demand prediction methods reported in the literature is hard because: (1) these methods were tested in different experiment design settings, e.g. different prediction windows and temporal resolutions, (2) the test datasets were different.
Consequently, some helpful, or obstructing, phenomena present in one dataset may be absent in another. Second, real datasets are often subject to cleaning procedures which are scarcely reported. Finally, the reported quality scores use different measures of error, e.g., MAPE or MSE. Hence, the reported numerical values of prediction quality should be taken with caution.

\section{Problem Formulation} \label{sec:formul}
Energy distribution and measurement in the DHS has a tree structure. In selected nodes, energy counters are installed. There are two types of energy counters, related to two types of energy usage: (1) hot water energy (HW) counters (e.g., used for washing, bathing, showering) and (2) heat energy (HE) counters (for heating buildings and adsorption cooling). A sum of the whole network energy usage is recorded in the tree root.
Prediction of the energy "sum" in the tree root is required to plan selling electricity on day-ahead markets. 

Past energy readings, exogenous weather factors, e.g., atmospheric temperature, humidity, and wind speed are known in equally spaced 1 hour resolution. A three-day weather forecast is also known with 1 hour resolution. Thus, these values are treated like time series. 
All the past data may be used to develop, i.e. train, a forecasting algorithm. It is required to forecast any counter readings in the next three days. Notice that our goal is slightly different than in many publications on DHS forecasting, because we need to forecast energy consumption for each counter (not only the "sum" counter) and for relatively short intervals.

Formally, given past energy readings of a certain counter
${\cal Y}=(Y_1,\dots,Y_t)$ ($Y_i$ are scalars), past values of exogenous variable values 
${\cal W}=(W_1,\dots,W_t)$ (where $W_i$ are vectors), and their forecast 
${\cal W}'=(W_{t+1},\dots,W_{t+72})$
it is required to forecast energy demands
${\cal F}=(Y_{t+1},\dots,Y_{t+72})$.

\section{Algorithms} \label{sec:alg}

\subsection{Dotzauer Method} \label{sec:Dotzauer}
In \cite{Dotz02} heat demand $Y_i$ for hour $i$ is modeled as a sum of two components:
\begin{equation}
Y_i=f(T_i)+g(i).\label{eq:Dotz1}
\end{equation}
The first component is a function of the atmospheric temperature $T_i$ at hour $i$. The second one is an array of corrections calculated for hours in a week. $f(T_i)$ is a piecewise linear function with five segments. 
$g(i)$ is referred to as {\em social component} because it accounts for customers habits.
In \cite{Dotz02} hourly corrections $g(i)$, segment levels and slopes in function $f$
were chosen by minimizing the squared error between the model and the past observations.
For a future hour $j$ and temperature forecast $T_j$, heat demand can be computed from (\ref{eq:Dotz1}).

\begin{figure}[t]
\setlength{\unitlength}{1cm}
\begin{picture}(12,5.5)
\put(-.6,0.3){\includegraphics[width=6.8cm]{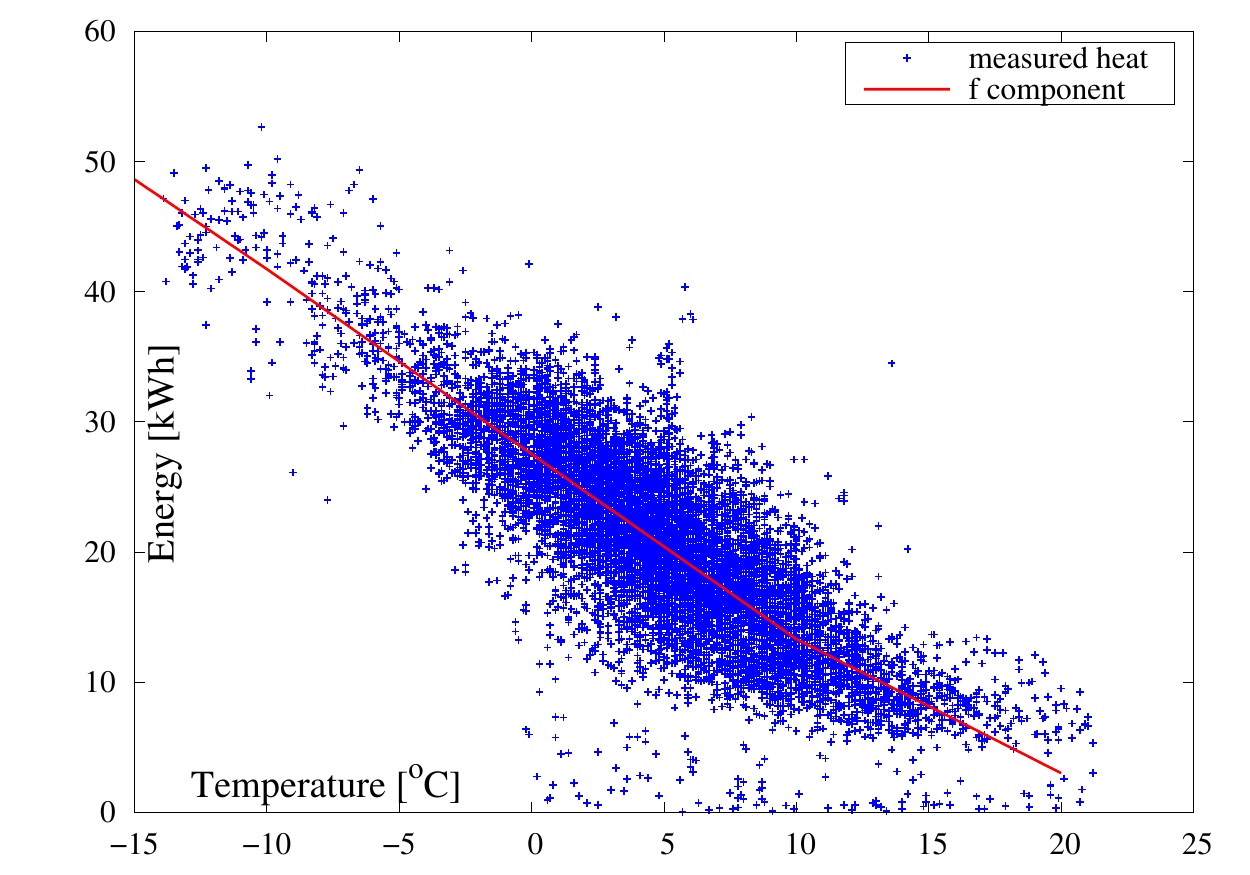}}
\put(5.7,0.3){\includegraphics[width=6.8cm]{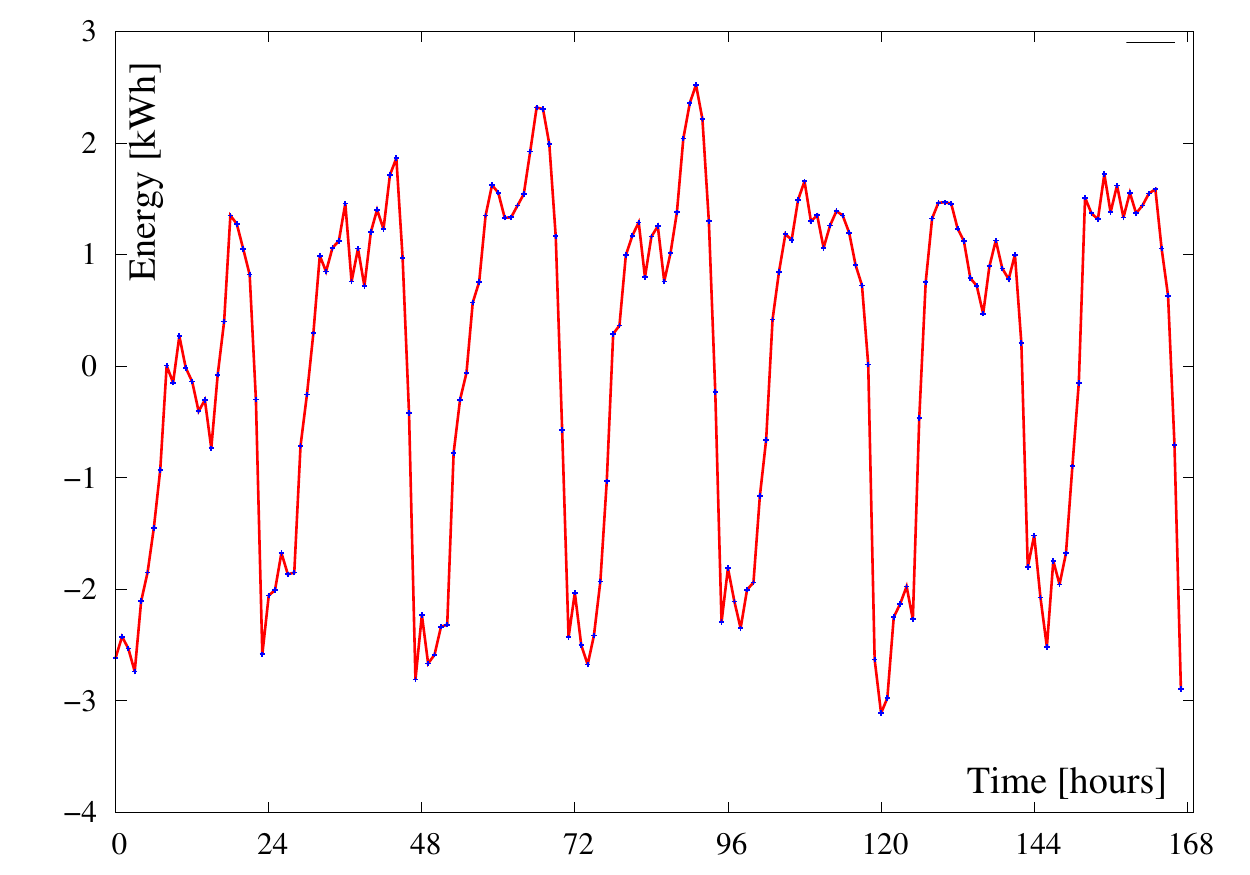}}
\put(3,-0.){a)}
\put(9,-0.){b)}
\end{picture} 
\caption{Examples of heat energy counter model equation (\ref{eq:Dotz1}) components.
a) Temperature component $f(T)$.
b) Social component $g(i)$.}
\label{fig:Dotz-g-f-exmp-heat}
\end{figure}

In our implementation also five segments were used in the atmospheric temperature model.
Temperature positions of the break-points in the piecewise linear function $f$
were chosen such that the past temperature readings were divided into five equicardinal subsets.
A shorthand notation DPLW (for Dotzauer method with Piecewise Linear temperature model and Weekly corrections) will be used to refer to this algorithm.
Example components $f$ and $g$ from (\ref{eq:Dotz1}) are shown in 
Fig.\ref{fig:Dotz-g-f-exmp-heat}. Let us observe that range of $g(i)$ is roughly one tenth of $f$ range in Fig.\ref{fig:Dotz-g-f-exmp-heat}. 

\subsection{Dotzauer Method Extensions} \label{sec:Dotz-ext}
In this section we propose new variants of Dotzauer method that differ in the 
social and temperature components construction.

\smallskip

\noindent
{\bf Social Component Models.}
Two versions of component $g(i)$ were studied:

\noindent
{\em Weekly corrections:}
Function $g(i)$ is a sequence of 168 hourly values for each hour $i$ of a week.
It was calculated as an average difference $Y_i-f(T_i)$ for all week hours in the training dataset.

\noindent
{\em Yearly corrections:}
$g(i)$ is calculated analogously, but for each hour of the year.

\smallskip

\noindent
{\bf Temperature Component Models.}
The following temperature component models were implemented:

\noindent
{\em Linear Temperature Model:}
$f(T_i)$ is linear, which is a simplification of DPLW.

\noindent
{\em Spline Temperature Model:}
$f(T_i)$ is a cubic spline with four internal knots, 
i.e. a function that is a sequence of five piecewise cubic polynomials 
and is twice continuously differentiable.
Temperature positions of the four knots in the spline were determined as in DPLW.
Similarly to DPLW, parameters of the spline polynomials were chosen to minimize 
the squared error between the model and the past observations.

\noindent
{\em Isotonic Regression Model:}
$f(T_i)$ is constructed by applying isotonic regression.
This method fits a free-form piecewise line of non-increasing heat energy consumption into a sequence of given temperature-heat points for minimum square error.
Pool adjacent violators algorithm is used for this purpose.

\noindent
{\em Multivariate Temperature Model:}
The temperature component is a multivariate linear function
of exogenous variables related to weather conditions.
The combinations of weather conditions applied are listed in Tab.\ref{tab:a0-4}
in appendix \ref{sec:exo-weather}.

\smallskip

\noindent
The following versions of the Dotzauer method were used in the further studies:\\
$\bullet$ Linear Temperature Model Weekly/Yearly Corrections (DLW/DLY),\\
$\bullet$ Piecewise Linear Temperature Weekly/Yearly Corrections (DPLW/DPLY),\\
$\bullet$ Spline Temperature Model Weekly/Yearly Corrections (DSW/DSY),\\
$\bullet$ Isotonic Temperature Model Weekly/Yearly Corrections (DIW/DIY),\\
$\bullet$ Multivariate Temperature Model Weekly/Yearly Corrections (DMW/DMY).

\subsection{W-Regressors}
W-regressors are a set of methods building independent linear-regression models for each hour of the prediction interval.
Since the linear regression is built on the past observations of the energy consumption, W-regressors can be considered as auto-regressive methods.

\smallskip


\noindent
{\bf WRNH}
builds independent prediction models for each hour of the week.
Consequently, to predict in  72-hour window, a subset of 72 out of 168 models will be used.
For each hour $i$ of the week a moving average of the last week energy consumption is calculated (168 samples).
In order to construct a linear regression fit for hour $i$ tuples 
$(a_j,W_j,Y_j)$ are used as input data points, where:
$a_j$ is energy consumption 168-hour moving average at hour $j$,
$W_j$ are weather conditions at hour $j$,
$Y_j$ is the actual energy consumption at hour $j$, 
for hours $j:j \bmod 168=i$ in the past.
We applied various combinations of atmospheric conditions in vector $W_j$. 
This resulted in five versions WRNH0,$\dots$,WRNH4 of this algorithm. 
Used weather-related components of $W_j$ are listed Tab.\ref{tab:a0-4}.
Tuples $(a_j,W_j,Y_j)$ are used to fit linear regression 
$Y_i=\overline{k_i}\times [a,W]^T +l_i$,
where $a$ is a moving average, $W$ is a vector of weather conditions,
$[a,W]$ is a vector of independent variables,
$\overline{k_i}$ is a vector of directional coefficients,
and $Y_i$ is the modeled energy consumption.
When predicting energy consumption for future hour $t+p$, for $p=1,\dots,72$, at the current hour $t$, 
the energy prediction is calculated as 
$Y_{t+p}=\overline{k}_{(t+p)\bmod 168}\times [a_t,W_{t+p}]^T+l_{(t+p)\bmod 168}$, 
where $a_t$ is the moving average of energy consumption at prediction moment $t$,
$W_{t+p}$ is a forecast of weather conditions for hour $t+p$.
The nickname of these methods comes W-Regressor No History (WRNH).

\smallskip


\noindent
{\bf WRWH.}
While predicting energy demand, the previous method assumed that at future hour $t+p$ 
moving average $a_t$ is used independently of the location of hour $t$ in the week.
An advantage is that only 168 models are needed.
A disadvantage is that the current hours $t$ may correspond to very different social conditions in a week.
It is reasonable to correlate the prediction for the future hour $t+p$ with the moving average 
for each particular hour of the week $t$.
Thus, WRWH regressors (short for W-Regresor With History) build $168 \times 72=12096$ linear models.
72 of them are used to build a 72-hour prediction at the current hour $t$.
In order to develop a model for future hour $i+p$, where $i=t \bmod 168$ 
were the hours of the week of the current moment $t$,
tuples $(a_j,W_j,Y_q)$ for $j:j\bmod 168=i$ and $q:q=j+p$, were used.
Future hour $t+p$ energy consumption is calculated as
$Y_{t+p}=\overline{k}_{(t\bmod 168),p}\times [a_t,W_{t+p}]^T+l_{(t\bmod 168),p}$,
where 
$\overline{k}_{i,p}$ is a vector of directional coefficients for the current hour of the week $i$ and energy consumption shifted $p$ hours into the future.
As in the previous case there are five versions WRWH0,\dots,WRWH4 of this method 
differing in the set of used weather features, cf. Tab.\ref{tab:a0-4} appendix \ref{sec:exo-weather}.

\subsection{Neural Networks}

\noindent
{\bf Feed-Forward Neural Network.}
Following the recent popularity of deep learning, we decided to also use a deep neural network
to calculate forecasts $\mathcal{F}$. 
We used a feed-forward neural network (FFNN for short) implemented in PyTorch.
Optuna, a hyperparameter optimization framework \cite{optuna_2019}, was applied 
to set up hyperparameters of the model. 
We randomly selected one of the counters, and optimized the following hyperparameters 
over the span of 500 trials:\\
-- number of hidden layers, considering the range from 1 to 5,\\
-- sizes of hidden layers, from the range 10 to 200 neurons,\\\
-- the dropout probability, from the range 0 to 0.5,\\
-- which past energy readings to use, for hourly back-shift $i\in\{0,1,\ldots,143\}$. \\
We used MSE as the optimization criterion for Optuna and arrived at the 
architecture presented in Tab.\ref{tab:deep-architecture}.
The indices $i$ of the past energy reading used as the inputs of FFNN are given in appendix \ref{sec:inp-ffnn}.
Beyond the past energy readings also all weather attributes mentioned in appendix \ref{sec:exo-weather},
for the current time $t$ were inputs to FFNN.
The network had 72 outputs to forecast for the 72 hours.

\begin{table}[t]
\caption{\label{tab:deep-architecture}%
Architecture of the FFNN. If not specified, the defaults of PyTorch  were used.}
\centering
\begin{tabular}{lll}
layer & component & class from \texttt{torch.nn} \\
\hline
Input & Batch normalization & \texttt{BatchNorm1d} \\
\hline
Hidden 1 & Linear $99\times 10$ & \texttt{Linear} \\
& Batch normalization & \texttt{BatchNorm1d} \\
& Leaky ReLU & \texttt{LeakyReLU} \\
& Dropout $p=0.406$ & \texttt{Dropout} \\
\hline
Hidden 2 & Linear $10\times 46$ & \texttt{Linear} \\
& Batch normalization & \texttt{BatchNorm1d} \\
& Leaky ReLU & \texttt{LeakyReLU} \\
& Dropout $p=0.406$ & \texttt{Dropout} \\
\hline
Output & Linear $46\times 72$ &  \texttt{Linear} \\
\hline
\end{tabular}
\end{table}

A separate neural network using the same architecture was trained for each counter. 
MSE was applied as the loss function and optimized using the Adam optimizer. 
Given the relatively small amount of training data, we employed the following techniques to avoid overfitting:\\
-- From the training set, we removed the most recent 10\% of observations and used them as the validation set to enable early stopping after 20 epochs without MSE improvement on the validation set. \\
-- We used the dropout with $p=0.406$ at the end of each hidden layer \cite{DBLP:journals/corr/abs-1207-0580}.\\
-- Within each hidden layer, we used batch normalization to avoid problems with exploding and vanishing gradients \cite{DBLP:conf/icml/IoffeS15}. 
Batch normalization in the input of the neural network was applied to ensure that the input data is normalized.\\

\noindent
{\bf Radial Basis Function Neural Network.}
As reported in Section \ref{sec:lietrature} neural networks with radial basis functions (RBFNN) were successful in energy usage prediction. 
To see how they fare with our data, we used the framework developed for FFNN, but used a single hidden layer consisting solely of RBF neurons. 
To decide on the number of neurons in this layer and the input features we used Optuna, as previously described, and arrived at 16 neurons.
The indices of past energy readings used as the inputs are shown in appendix \ref{sec:inp-rbfnn}.
For the output layer, we used a linear layer with 72 neurons, optimized using the same procedure as for FFNN.

\bigskip 

\noindent
A moving average of the last 100 hours was used as a reference prediction algorithm.
It is denoted as algorithm C-100.

\section{DHS Dataset} \label{sec:dataset}
The test dataset comprises heat measurements from 28 hot water and 83 heat energy counters in the DHS operated by Kogeneracja Zach\'od, in a town of about 30000 inhabitants.
The records cover the period since the 1st of September 2015 at 00:00 (start of day) until 28th of February 2019 at 23:00 (the last hour of a day). 
The dataset includes the following values: measurement timestamp, 
measured energy consumption, weather conditions, type of day, season (cf. \cite{KZData}). 
The measurements were collected for billing purposes, rather than for
heat demand prediction and optimization, so they 
have several drawbacks from the datamining perspective:\\
$\bullet$ Since clients join and leave the DHS, the covered interval is different for particular counters.\\
$\bullet$ A measurement of zero energy consumption may mean that either a customer left the DHS, malfunctioning of the counter, or maintenance time. Hence, there is a large fraction of 0-measurements, which need special handling.\\
$\bullet$ There are intervals when no energy use is recorded for the whole DHS.\\
$\bullet$ The majority of heat energy consumption happens between the beginning 
of October and the beginning of May.\\
$\bullet$ It is known that to economize on energy, hot water supply was occasionally shut down at night hours by property owners. This process had random nature and it is not known to which counters and on what days it happened.\\
$\bullet$ In a large part of weather records only 0 values are present. Since such values (e.g., temperature in $^\circ$C, wind speed) are admissible, it was not possible to distinguish missing and correct data.

In order to deal with the missing and dubious data the following cleaning procedures were applied:\\
$\bullet$
zero energy readings are valid if preceded and succeeded by a non-zero reading (to eliminate zeros before switching a counter on and after switching it off);\\
$\bullet$
at the considered time, at least one counter has non-zero recording
(this eliminates, e.g., maintenance periods and intervals where data
was missing globally);\\
$\bullet$ 
only readings with non-zero humidity are accepted (this eliminates readings with dubious weather recordings).

It was observed that customers significantly differ because the ratios of the smallest to the largest average hourly energy usage in kWh were 7:57 for HW and 4:125 for HE counters. 
A strong correlation between atmospheric temperature and HE recordings was observed because correlation coefficient was in range $[-0.922,-0.706]$, depending on the counter. For hot water correlation coefficient was in range $[-0.164,0]$.

\section{Experimental analysis of prediction algorithms} \label{sec:eval}
\begin{figure}[t]
\setlength{\unitlength}{1cm}
\begin{picture}(12,5.5)
\put(-.6,0.3){\includegraphics[width=6.6cm]{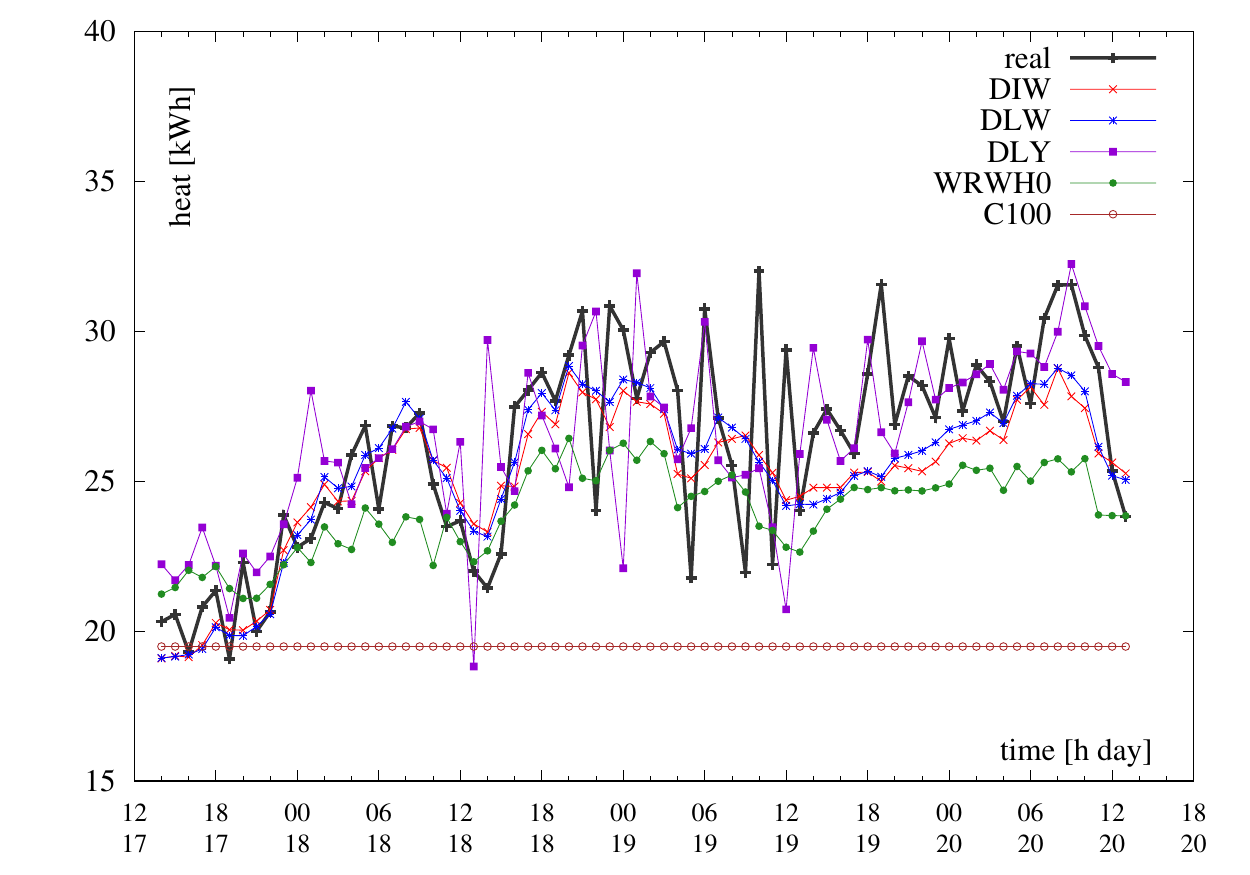}}%
\put(5.7,0.3){\includegraphics[width=6.6cm]{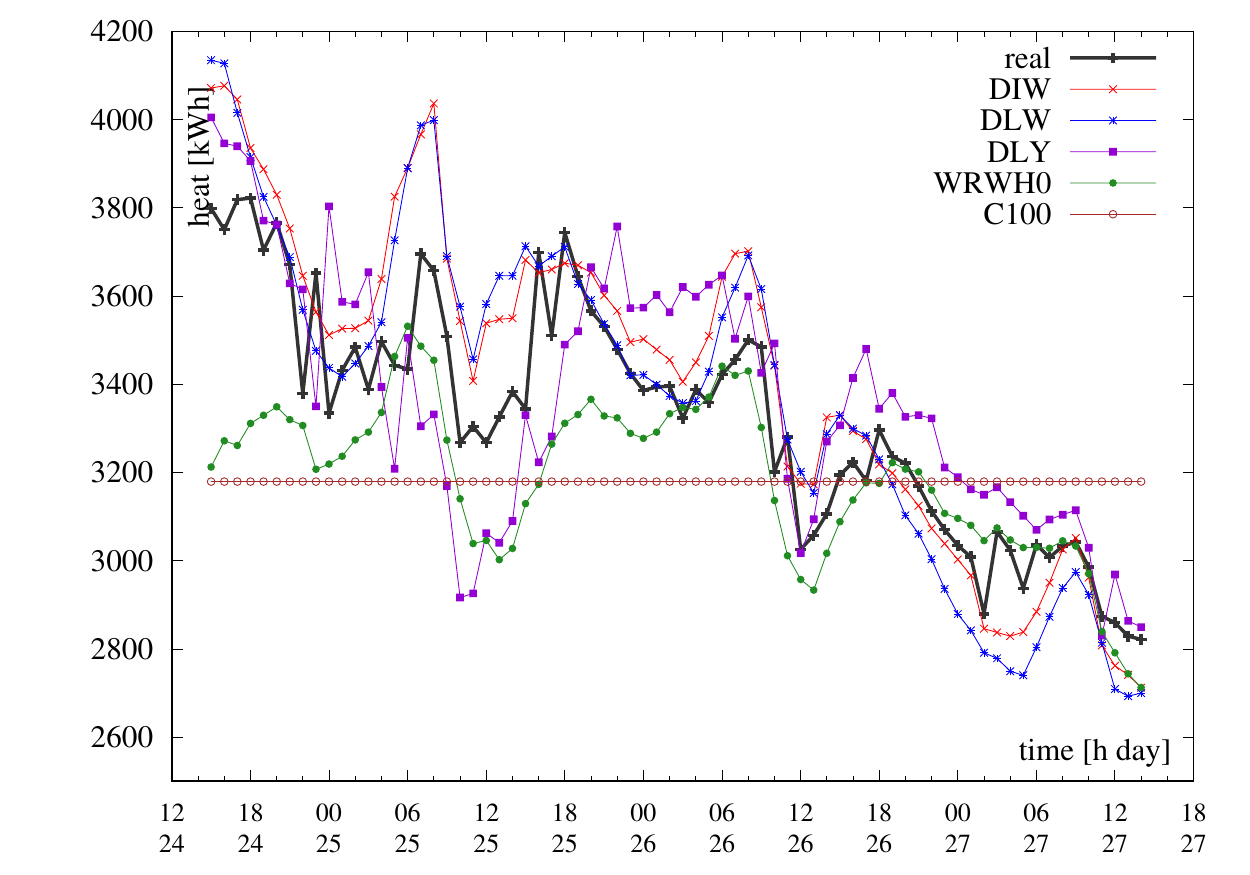}}%
\put(3,-0.){a)}
\put(9,-0.){b)}
\end{picture} 
\caption{Examples of heat demand prediction vs time in 72-hour window.
a) One of the heat counters.
b) "Sum" counter.}
\label{fig:ValvsTi}
\end{figure} 

Prediction algorithms were subjected to the following test framework:
Data from interval 2016-01-01 00:00:00 until 2017-12-31 23:00:00
was used as the training dataset.
Interval from 2019-01-02 00:00:00 until 2019-02-25 23:00:00 was used for testing.
For other combinations of training and testing intervals similar results were obtained so we do not report them.
In the testing interval, each of the algorithms calculated a 72-hour 
energy consumption forecast starting at each hour of the testing interval
(1320$\times$72 prediction points).
The training and testing were conducted for each energy counter.
All the codes were written in Python 3.7.4 and tested on a PC with Windows 10 
and Intel i7-8550U CPU @ 1.80GHz (no GPU acceleration).

\subsection{Basic evaluation}

Representative examples of predictions over 72-hour interval 
are visualized in Fig.\ref{fig:ValvsTi}.
Only a subset of nondominated methods is shown to avoid cluttering the picture
(the nondominated methods will be explained in the next section).
Fig.\ref{fig:ValvsTi}a presents results for one of the counters,
while Fig.\ref{fig:ValvsTi}b for the "sum" counter.
Note that the vertical axes start at values greater than 0.
It can be verified that visually these algorithms follow the real energy consumption.

Quality of the predictions aggregated over all 1320$\times$72 prediction points 
are presented quantitatively in Tab.\ref{tab-Sum} and  Tab.\ref{tab-best-dev}.
In Tab.\ref{tab-Sum}, 8 best algorithms for the "sum" counter with respect to
MAPE and MSE measures are shown.
It can be seen that RBFNN and simple Dotzauer model variants with weekly corrections provided the best predictions of the "sum" of consumed energy for MAPE.
For MSE quality measure RBFNN, FFNN are the best algorithm whereas 
C-100 and DLW (which is the best Dotzauer-like algorithm)
have MSE three times worse than FFNN (not shown in Tab.\ref{tab-Sum}).
Thus, RBFNN, FFNN are the best for "sum" counter.
In Tab.\ref{tab-best-dev} the best MAPE obtained on any heat counter
is given for each prediction algorithm.
In a sense, this table shows the limitations of the prediction algorithms
on the available data.
Surprisingly, moving average (C-100) used as a reference algorithm 
is capable of obtaining good results in comparison with other methods.
Only WRNH0 and WRWH0 were able to compete.
Yet, the distribution of algorithm MAPEs across counters is more complex,
and the fact that C-100 is good on one counter does not imply that it 
is good overall.
This will be further discussed in the next section.
Furthermore, such a result rises reservations about quality of the data.

\begin{table}[t]
\caption{8 best algorithms on "Sum" counter}\label{tab-Sum}%
\centering
{\scriptsize
\begin{tabular}{|l|l|l|l|l|l|l|l|l|l|l|l|l|l|}
\hline
\multicolumn{9}{|c|}{MAPE}\\
\hline
method & RBFNN  & DLW & DMW  & DPLW & DIW & WRWH0 & FFNN & DSW \\
\hline
MAPE [\%] & 16.45 & 16.49 & 16.49 & 16.96  & 17.1 & 17.7 & 18.43 & 18.45 \\ 
\hline
\multicolumn{9}{|c|}{MSE}\\
\hline
algo. & FFNN & RBFNN & WRWH0 & WRWH1 & WRNH0 & WRNH1 & WRNH2 & WRWH2 \\
\hline
MSE & 217884 & 237987 & 437482 & 452190 & 463599 & 477858 & 568191 & 575916 \\
\hline
\end{tabular}%
}
\end{table}

\begin{table}[t]
\caption{Best results for the prediction algorithms}\label{tab-best-dev}%
\centering
{\scriptsize
\begin{tabular}{|l|l|l|l|l|l|l|l|l|l|l|l|l|l|}
\hline
method      &C-100	&DIW	&DIY	&DLW	&DLY	&DMW	&DMY	&DPLW\\
\hline
MAPE [\%]   &8.3	&10.2	&11.0	&10.0	&11.5	&10.0	&11.5	&10.1\\
\hline
method      &DPLY	&DSW	&DSY	&FFNN	&RBFNN	&WRNH0	&WRNH1	&WRNH2\\
\hline
MAPE [\%]   &11.0	&10.0	&11.5	&9.8	&16.5	&8.1	&9.6	&15.5\\
\hline
method      &WRNH3	&WRNH4	&WRWH0	&WRWH1	&WRWH2	&WRWH3	&WRWH4&\\
\hline
MAPE [\%]   &17.5	&8.4	&7.9	&9.7	&17.2	&18.1	&8.5&\\
\hline
\end{tabular}%
}
\end{table}

\subsection{Time-Quality Trade-off}

\begin{figure}[t]
\setlength{\unitlength}{1cm}
\begin{picture}(12,5.5)
\put(-.5,0.3){\includegraphics[width=6.6cm]{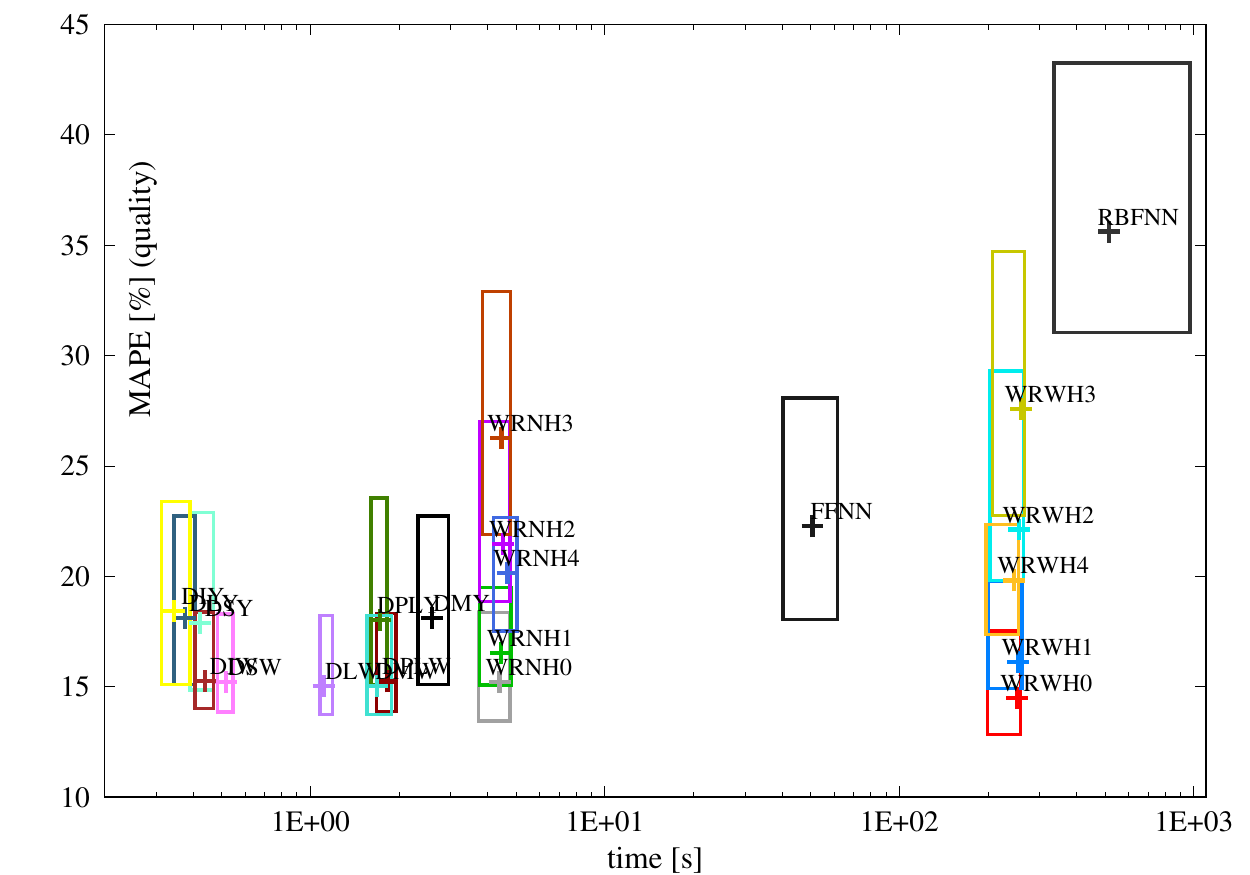}}
\put(5.8,0.3){\includegraphics[width=6.6cm]{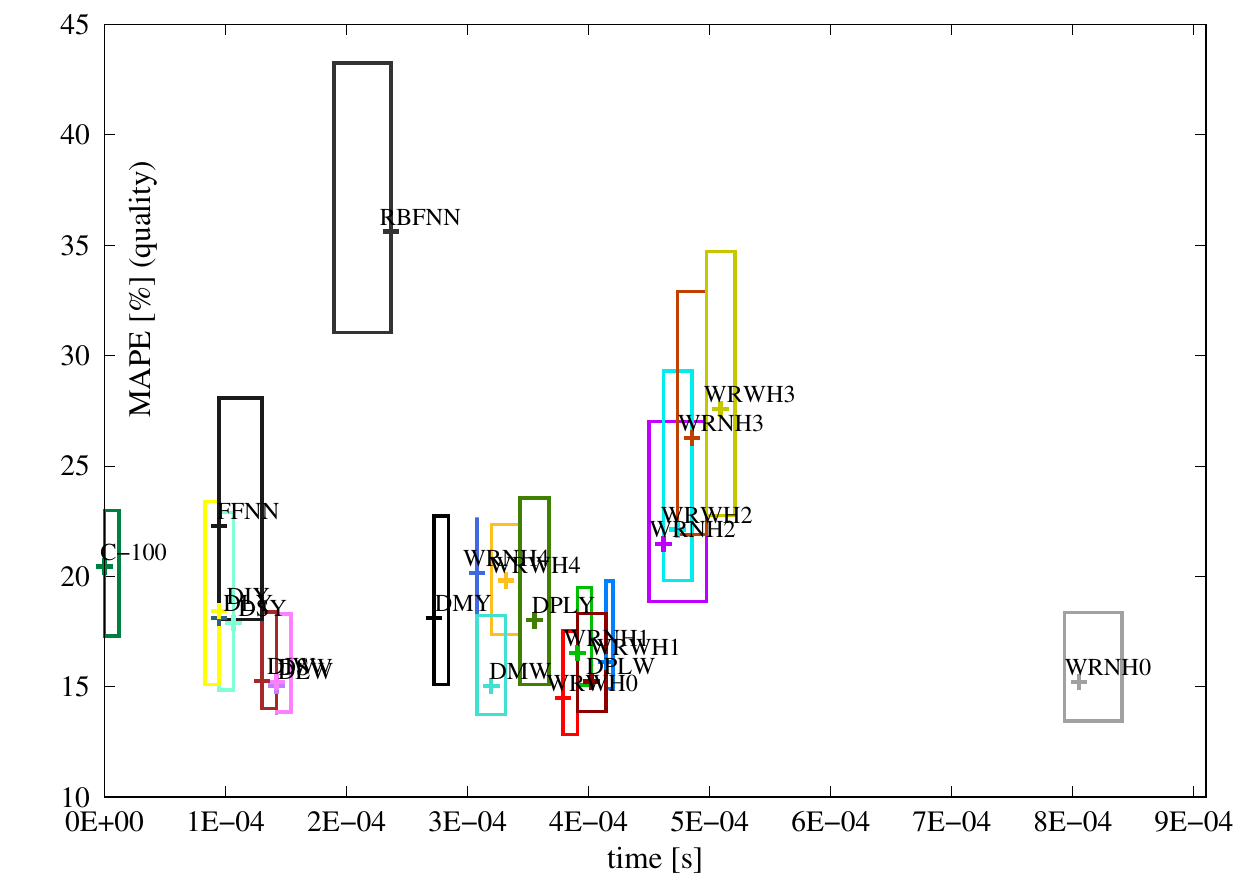}}
\put(3,-0.){a)}
\put(9,-0.){b)}
\end{picture} 
\caption{Prediction quality (MAPE) vs computational cost.
a) Prediction quality vs training time.
b) Prediction quality vs predicting time.}
\label{fig:QvsT}
\end{figure} 

Time performance of the prediction algorithms is important when these
algorithms are supposed to be used in low-power IoT or embedded devices.
In Fig.\ref{fig:QvsT} run-time vs quality trade-off is shown.
On the horizontal axis run-time is shown, on the vertical axis prediction quality
(MAPE) is presented.
Results for each algorithm are presented as interquartile boxes aggregated over all counters. 
That is, a box for each algorithm spans between Q1 and Q3 in time and prediction quality.
Median of quality and run-time is also marked.
This way of visualizing mutual algorithm performance has three-fold advantages:
1) it is possible to recognize algorithm differences with respect to prediction quality,
2) differences in time efficiency are visible,
3) it is possible to analyze how these algorithms trade run-time for prediction error.
In Fig.\ref{fig:QvsT}a training time is shown, in Fig.\ref{fig:QvsT}b time of calculating
a single 72-hour forecast is shown.
In Fig.\ref{fig:QvsT}a moving average C-100 algorithm is not presented because it is not trained.

For the training time the algorithms form clusters resulting from 
the computational complexity of the training procedure.
RBFNN are the slowest methods, then WRWH, FFNN, WRNH, and Dotzauer algorithms are gradually faster.
The algorithms with isotonic and spline temperature components are the fastest.
WRNH and WRWH algorithms have very tight run-time distributions, while WRWH methods are
apparently slower because 72 times more linear models must be computed.
As far as training these algorithms in low-power computers, Dotzauer-like and WRNH algorithms 
are feasible choices, whereas using RBFNN or WRWH algorithms seems disputable. 
Conversely, calculating one 72-hour prediction is far less costly
computationally and all the considered algorithms managed it in less than 1ms.

Considering prediction quality, most of the interquartile ranges overlap,
so it is rather hard to draw sharp conclusions on algorithm superiority,
but still, some tendencies can be observed.
The RBFNN accuracy is the worst, which can be attributed to great irregularity
of energy readings.
In WRNH and WRWH methods various weather attributes were used.
WRNH4 and WRWH4 can be taken as a reference because in this pair
no weather data was used (see Tab.\ref{tab:a0-4}).
Algorithms WRNH0, WRWH0 using only atmospheric temperature and day length 
are the best among WRNH/WRWH methods.
Using an extensive set of weather attributes, impedes rather than helps
obtaining good energy predictions.
Among Dotzauer-like algorithms the weekly social corrections are generally better
than yearly corrections, which is not surprising because there were only two years 
in the training dataset, and the weekly pattern in energy usage is strong (cf.
Fig.\ref{fig:Dotz-g-f-exmp-heat}).
It can be also verified in Fig.\ref{fig:QvsT} that C-100 method is not
the best overall and it has apparent competitors (e.g. WRWH0).
This contrasts with the result in Tab.\ref{tab-best-dev}, confirming that 
accuracy distributions of the algorithms across energy counters are different.

Fig.\ref{fig:QvsT} allows also to identify algorithms that are nondominated.
We used median run-time and median accuracy as indicators of
algorithm position on the run-time vs quality space.
The set of nondominated methods comprises the algorithms for which no other
algorithm has both better accuracy and run-time.
For better exposition, the nondominated algorithms for MAPE and MSE quality 
measures are collected in Tab.\ref{tab-pareto} in Section \ref{sec:app:nondominated}.
Depending on the setting, C-100, various versions of Dotzauer method and WRWH0
are chosen to the Pareto-front.
In all settings C-100 is selected as the fastest method.
Among the algorithms that are present in all Pareto-front settings
DLY and DIW are the second and third fastest algorithms, while
DLW and WRWH0 are the slowest pair.
These methods can be recommended as candidates for low-power computer systems.

\section{Conclusions} \label{sec:Concl}
In this paper algorithms for DHS energy consumption prediction were analyzed.
The goal of this study was to verify the utility of a set of algorithms in predicting
energy consumption every hour, in 72-hour intervals, both total 
and for each energy counter separately.
Another aspect, important in the use on low-power devices,
was the computational complexity of the algorithms.
It turned out that simple methods offer the best run-time quality trade-off.
Machine learning methods, FFNN, RBFNN, have low potential for training on low-power devices.
However, a more detailed inspection revealed that accuracy of energy consumption
prediction depends very much on the energy counter and accuracy measure.
For example, RBFNN, FFNN are the best at total energy prediction.
Future work may involve examining other machine learning methods from the literature,
comparing them with the algorithms presented here in the run-time quality setting.
Online updating of the models with new data should be also examined.
\\
\\
\textbf{Acknowledgements}. This research was partially supported by Kogeneracja Zach\'od S.A. 

\bibliographystyle{abbrv}
\bibliography{KogezArt-bib}

\appendix

\section{Indices of past energy readings used as FFNN input}
\label{sec:inp-ffnn}

Below shift $i$ into the past is given in hours with respect to the current hour.
For example, shift 0 is the last (the current) energy reading,
1 is for the previous energy reading.
The indices follow:

\noindent
{\small
0, 1, 2, 6, 7, 8, 9, 11, 12, 17, 19, 20, 21, 22, 23, 25, 26, 28, 29, 30, 32, 33, 35, 36, 37, 39, 40,
41, 42, 44, 46, 47, 48, 50, 53, 54, 56, 59, 60, 62, 67, 69, 71, 74, 75, 77, 82, 84, 85, 86, 92, 94, 
97, 98, 100, 103, 105, 106, 107, 109, 110, 112, 114, 116, 117, 118, 119, 121, 122, 125, 126, 127, 128, 
129, 130, 131, 132, 134, 135, 136, 137, 139, 140, 142,143.
}

\section{Indices of past energy readings used as RBFNN input}
\label{sec:inp-rbfnn}

Below shift $i$ into the past is given in hours with respect to the current hour.
For example, 1 is for the previous energy reading.
The indices follow:

\noindent
{\small
1, 2, 4, 5, 10, 14, 15, 19, 21, 23, 27, 28, 29, 41, 44, 45, 46, 47, 48, 51, 56, 57, 58, 60, 
65, 67, 72, 73, 75, 76, 79, 81, 86, 87, 90, 91, 93, 96, 104, 112, 116, 121, 124, 128, 131, 
132, 135, 139,140, 142
}

\section{Exogenous  variables used in WRNH, WRWH, DMW, DMY algorithms}
\label{sec:exo-weather}

\begin{table}%
\centering
\caption{\label{tab:a0-4}
Notation:
$T$ - atmospheric temperature, $DL$ - day length in hours, 
$DT$ - day type (1:Monday -- Thursday, 2:Friday, 3:Saturday, 4:Sunday),
$V$ - wind speed in m/s, $\sqrt{V}$ - square root of wind speed, 
$TV$ - product of temperature $T$ and wind speed ${V}$,
$T\sqrt{V}$ - product of temperature $T$ and $\sqrt{V}$,
$pY$ - season of the year (spring,$\dots$,winter),
$Oc$ - overcast in oktas,
$H$ - humidity.}
{\footnotesize
\begin{tabular}{|c|c|c|c|c|c|c|c|c|c|c|c|c|c|c|}
\hline
algorithm version& $T$ & $DL$ & $DT$  & $V$ & $\sqrt{V}$ & $TV$   & $T\sqrt{V}$ & $pY$ & $Oc$ & $H$ \\
\hline
WRNH0/WRWH0     & *   & *    &       &     &         &             &    &      &  &\\
\hline
WRNH1/WRWH1     & *   & *    &       &     &  *      &             & *   &      &  &\\
\hline
WRNH2/WRWH2     & *   & *    &       &  *  &  *      & *           & *  & *    &  &\\
\hline
WRNH3/WRWH3     & *   & *    & *     & *   &  *      & *           & *  & *    &* &*\\
\hline
WRNH4/WRWH4     &     &      &       &     &         &             &    &      &   &\\
\hline
DMW/DMY         & *   & *    & *     & *   &  *      & *           & *  & *    &* &*\\
\hline
\end{tabular}%
}
\end{table}%

\vspace*{-0.3cm}

\section{Nondominated algorithms}
\label{sec:app:nondominated}

\vspace*{-0.5cm}
\begin{table}[ht]
\centering
\caption{Nondominated algorithms}\label{tab-pareto}%
{\footnotesize
\begin{tabular}{|l|l|l|l|l|l|l|l|l|l|}
\hline
\multicolumn{8}{|c|}{MSE - training}\\
\hline
algorithm   & C-100 & DIY   & DLY   & DIW   & DLW  & DPLW  & WRWH0 \\
\hline
MSE         & 31.3  & 24.2  & 23.2  & 18.5  & 18.04 & 18.03 & 16.7\\
\hline
time [s]    & 0		& 0.344 & 0.375 & 0.438 & 1.11 & 1.83  & 251.2 \\
\hline
\multicolumn{8}{|c|}{MSE - predicting}\\
\hline
time [$\mu$s]& 11.8  & ~ --  & 94.7  & 130  & 142   & ~ -- & 379\\
\hline
\end{tabular}%

\bigskip

\begin{tabular}{|l|l|l|l|l|l|l|l|l|l|}
\hline
\multicolumn{9}{|c|}{MAPE - training}\\
\hline
algorithm & C-100 & DIY   & DLY   & DSY   & DIW   & DSW   & DLW  & WRWH0 \\
\hline
MAPE [\%] & 20.5  & 18.4  & 18.1  & 17.9  & 15.3  & 15.2  & 15.0 & 14.5 \\
\hline
time [s]  & 0	  & 0.344 & 0.375 & 0.422 & 0.438 & 0.516 & 1.11 & 251.2 \\
\hline
\multicolumn{9}{|c|}{MAPE - predicting}\\
\hline
time [$\mu$s]& 11.8  &~ --  & 94.7 & 107   & 130   &~ --  & 142  & 379 \\
\hline
\end{tabular}%
\smallskip
}
\end{table}%

\end{document}